\documentclass[11pt,letterpaper]{article}
\usepackage{cogsys}
\usepackage[T1]{fontenc}
\usepackage{times}
\usepackage[pdftex]{graphicx} 

\usepackage{natbib}
\usepackage{hyperref}
\hypersetup{
  colorlinks   = true, 
  urlcolor     = blue,
  linkcolor    = blue, 
  citecolor   = blue 
}
\cogsysheading{11}{2021}{1--**}{9/2021}{11/2021}

\begin{document} 

\ShortHeadings{Lensing Machines}
              {K.Dinakar and H.Lieberman}

\title{Lensing Machines: Representing Perspective in Latent Variable Models}
 
\author{Karthik Dinakar\textsuperscript{1,2}}{kdinakar@bwh.harvard.edu}
\author{Henry Lieberman\textsuperscript{1}}{lieber@media.mit.edu}
\address{\textsuperscript{1}Massachusetts Institute of Technology }
\address{\textsuperscript{2}Brigham \& Women's Hospital and Harvard University }
\vskip 0.2in
 
\begin{abstract}Many datasets represent a combination of several viewpoints -- different ways of looking at the same data that lead to different generalizations. For example, a corpus with examples generated by different people may be mixtures of many perspectives and can be viewed with different perspectives by others. It isn't always possible to represent the viewpoints by a clean separation, in advance, of examples representing each viewpoint and train a separate model for each viewpoint. We introduce \textbf{\textit{lensing}}, a mixed-initiative technique to \textbf{\textit{(1)}} extract '\textbf{\textit{lenses}}' or mappings between machine-learned representations and perspectives of human experts, and to \textbf{\textit{(2)}}  generate \textbf{\textit{'lensed'}} models that afford multiple perspectives of the same dataset. We apply lensing for two classes of latent variable models (a) a mixed-membership model and (b) a matrix factorization model in the context of two mental health applications, and we capture and imbue the perspectives of clinical psychologists into these models. Our work shows the benefits of the machine learning practitioner formally incorporating the perspective of a knowledgeable domain expert into their models rather than estimating unlensed models themselves in isolation.
\end{abstract}

\section{Introduction}
The importance of perspectives in the science of machines attempting to amplify human endeavors is captured succinctly by the famous quote \textit{'\textbf{A point of view is worth 80 IQ points}'} by Alan Kay \citep{kay1990point}. Perspectives or points of view -- critiquing and interpreting a corpus of work is central in the social sciences \citep{1976}, the design of user interfaces \citep{holtzblatt1993contextual}, and principles governing mixed-initiative interfaces \citep{horvitz1999principles}. Generalization in both generative and discriminative machine learning is venerated as one of the most important goals informing methodology and parameter estimation. Models are trained using data consisting of concrete instances, with care placed on sampling, regularization and hyperparameter optimization, towards estimating 'generalized' models that can be used for classification, prediction, or other tasks.

\begin{figure}[ht]
\begin{center}
\includegraphics[width=0.60\linewidth]{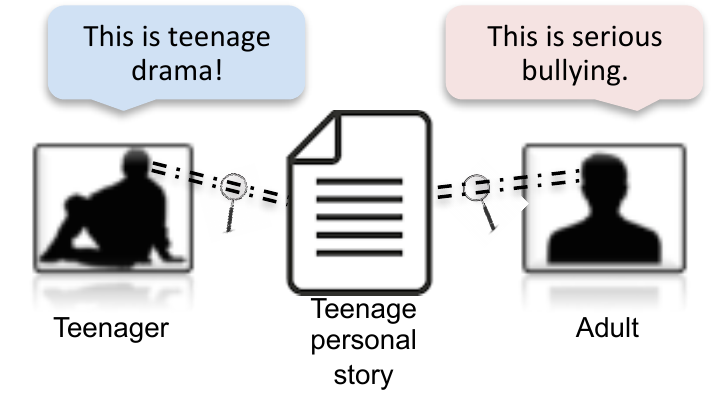}
\end{center}
\caption{Imagine a corpus of personal stories of hardship and distress by teenagers. A single story in the corpus where an aggressive interaction between two teenagers might be viewed as just 'mere drama' by teenagers themselves, but as 'serious bullying' by an adult, two different perspectives or 'lens' of the same story.}
\label{fig:perspectives}
\end{figure}

Traditional machine learning requires a dataset that annotates each example with a' label', usually single keywords per label. Single keyword labels don't always capture a nuanced understanding of the data, where different perspectives can govern the understanding of the data itself.

We'll give two examples. First, \textbf{(A)}: a corpus of stories by anguished and suicidal teenagers, In this dataset, a teenage interaction about 'bullying at home' and 'bullying at school' can be interpreted differently by a teenager and by an adult. 'Bullying at home' might be seen viewed by the adult as abuse, but considered by the teenager as 'drama', or vice versa. Asking the adult and the teenager to assign a single word or a topic to this story, or answer multiple-choice questions, is a problematic cognitive task -- the subjects' idea of a topic may not correspond well to the researchers' or the algorithms' representation of the story, documented extensively in the social sciences that investigates this phenomenon \citep{marwick2011drama}.

Dataset \textbf{(B)} consists of millions of posts on a mental health social support network website. It is a sparse matrix denoting users as rows and breadcrumb behaviors (such as clicks and browsing session information) as columns.

\begin{figure*}[ht]
\begin{center}
\includegraphics[width=8.50\linewidth]{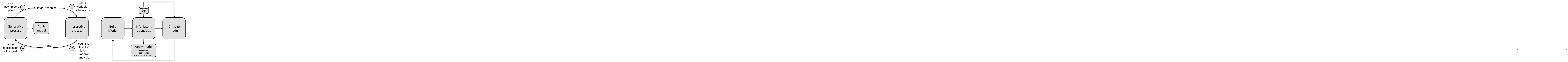}
\end{center}
\caption{The above figure shows the lensing process(on the left) and Box's loop (on the right). In lensing, (1) the generative model building process begins without separation of perspectives (or lens) from the training data in advance; (2) latent variables estimated as part of the generative process are then fed to the human in the loop; (3) who begins an interpretive process to interact with a cognitive task to assign semantic meaning to machine-generated latent variable distributions; (4) the assignment of meanings generates a vector of labels or 'lens' that is then merged with the priors of the generative task in the next iteration. Continually iterating steps (1) through (4) generates a model that captures that lens of the human in the loop, which can then be applied for purposes of classification, prediction and summarization etc. This is similar to Box's loop (on the right), where one can imagine model criticism as an exercise in mapping machine-generated latent variables to those that make semantic sense to the human in the loop. }
\label{fig:lensing}
\end{figure*}

In dataset \textbf{B}, say an unsupervised algorithm clusters users of the social network based on their online behavior, and these clusters are to be analyzed or annotated by a clinical therapist and by a user of the website. The annotation exercise by the therapist is informed by her or his perspective of the user behaviors manifest in each cluster that may not be the same as that of a distressed user needing emotional support. 

In the above two examples, the parties are simultaneously exploring and analyzing the datasets. This presents us two challenges: \textbf{(1)} how might we extract different perspectives of a dataset without reducing it to simply the assigning of single keyword labels?, and \textbf{(2)} how might we imbue these perspectives towards training multiple models that represent them. The paper is organized as follows: first, we define the terms lens (a perspective or representation of a dataset) and lensing (the process of using the lens to generate models that represent them); second, we list two principles governing the extraction of the lenses; third, we discuss the evaluation of the lensed models via model criticism (or analysis by synthesis) with Box's loop; lastly, we discuss three specific examples of lensing. Our work contributes to the growing area of human-in-the-loop computation in interactive machine learning.

\section{Lensing}
Lensing is our way of representing viewpoint or perspective in latent variable generative modeling.

For example, consider topic modeling. The most straightforward representation of a topic is a single keyword. Annotation tasks for supervised learning ask a human annotator to assign a keyword to a particular example. The annotator is asked a question like, "What is the topic of this news article"? or "What is shown in this picture?" The answers to these questions are aggregated over many examples, and perhaps many annotators, to form a corpus upon which the machine learning algorithm operates.

While these may seem like simple questions, it is not always a straightforward cognitive task for human annotators to answer. Different annotators may have a different understanding of what the keyword means or what the example is about. If forced to choose between a given list of alternatives, none may fully describe what the annotator thinks, or there may be "in-between" cases.

The conventional view on machine learning treats these as sources of error, to be minimized by the statistical power of large corpora. But they may not just represent random error. These variations may
systematically represent different points of view arising from personal circumstances, culture, linguistic differences, and other sources.  It may be imperative, as we shall see, that these points of view be understood. It may be necessary to be able to view a corpus according to diverse viewpoints.

Generative machine learning models, such as LDA, take a slightly more sophisticated view of a topic. Rather than a single word, they view a topic as a probability distribution over the words in the vocabulary. They view a document as a probability distribution over the topics so represented.

While this is a more sophisticated representation, it still may not correspond well to the understanding of the human annotator. So we are proposing to add a \textbf{\textit{human-in-the-loop}} step to the machine learning process. The purpose is to verify that the machine learning process faithfully captures human understanding by illuminating the correspondence between the machine's representation and the person's.

First, we run a conventional machine learning algorithm such as LDA, resulting in, for each example, a probability distribution over topics. Then we present to a person whose role we'll call the \textbf{\textit{informant}}, the original example, and/or some representation of the topic (a list of weighted keywords). We ask the informant an open-ended question, "What do you think this means?", and they can answer in unrestricted natural language, explaining any
subtleties that the example may hold.

\begin{figure*}[ht]
\begin{center}
\includegraphics[width=10.00\linewidth]{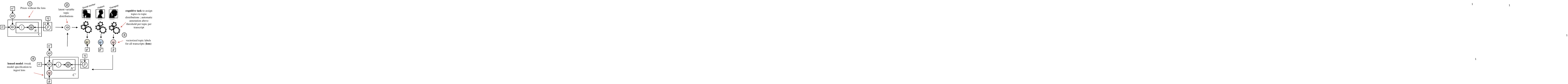}
\end{center}
\caption{\textbf{Lensing for crisis counseling:} Steps 1 through 4 relate directly to the steps outlined in figure \ref{fig:lensing}. The lens of the therapist in unknown initially. We begin with a simple hyper-parameterized model of LDA in step 1; the latent topic distributions are then presented to the therapist in step 2; in step 3, the therapist engages in a cognitive task where she generates complete sentences from topic distributions to assign a semantic and salient topic to each distribution, and these are folded into the corpus for the next iteration; the transcripts above with topic proportions above an assigned threshold are now given a set of therapist labels, which is a matrix of per-transcript labels for the entire corpus; step 4 tweaks the model specification to ingest this therapist lens towards a lensed model with topic and word coherence that aligns with the perspective of the therapist. There can be any lenses -- for example, if we replaced the therapist with a social worker or even a patient.} 
\label{fig:topicModelLensing}
\end{figure*}

We then can convert the answer back to a representation suitable for the machine learning algorithm (again, in the LDA case, a distribution over topics), and feed it back to the next round of the machine learning process. For example, in one application, we ask informants to construct a sentence that describes the example, using the most salient words in the topic.

From this, we construct a \textbf{\textit{lens}}, which is a mapping between latent variable distributions estimated as part of a general model building process, to semantic descriptions assigned to these distributions from the perspective of the informant. The lens is represented as a matrix of vectors.

We'll use the term \textbf{\textit{lensing}} to refer to the two-step process of, first, constructing the lens, and second, applying the lens to criticize the model specification. Our hope is that the latent variable distribution of the lensed model may make more semantic sense given the informant's perspective than the original vanilla representation.

In many applications, the informant's opinion may be more valuable, in some sense, than the original annotators. The informant may be a domain expert with more knowledge than the annotators and more valuable time. The informant may have personal experience, relevant context, or a unique point of view that wasn't available to the original annotators (or may not even be represented in the original corpus). Lensing provides a way to leverage the informant's opinion and the labor of the original annotators' labor without requiring the informant to annotate the entire original corpus. We now discuss two specific applications where lensing can be used.

\begin{figure*}[ht]
\begin{center}
\includegraphics[width=1.00\linewidth]{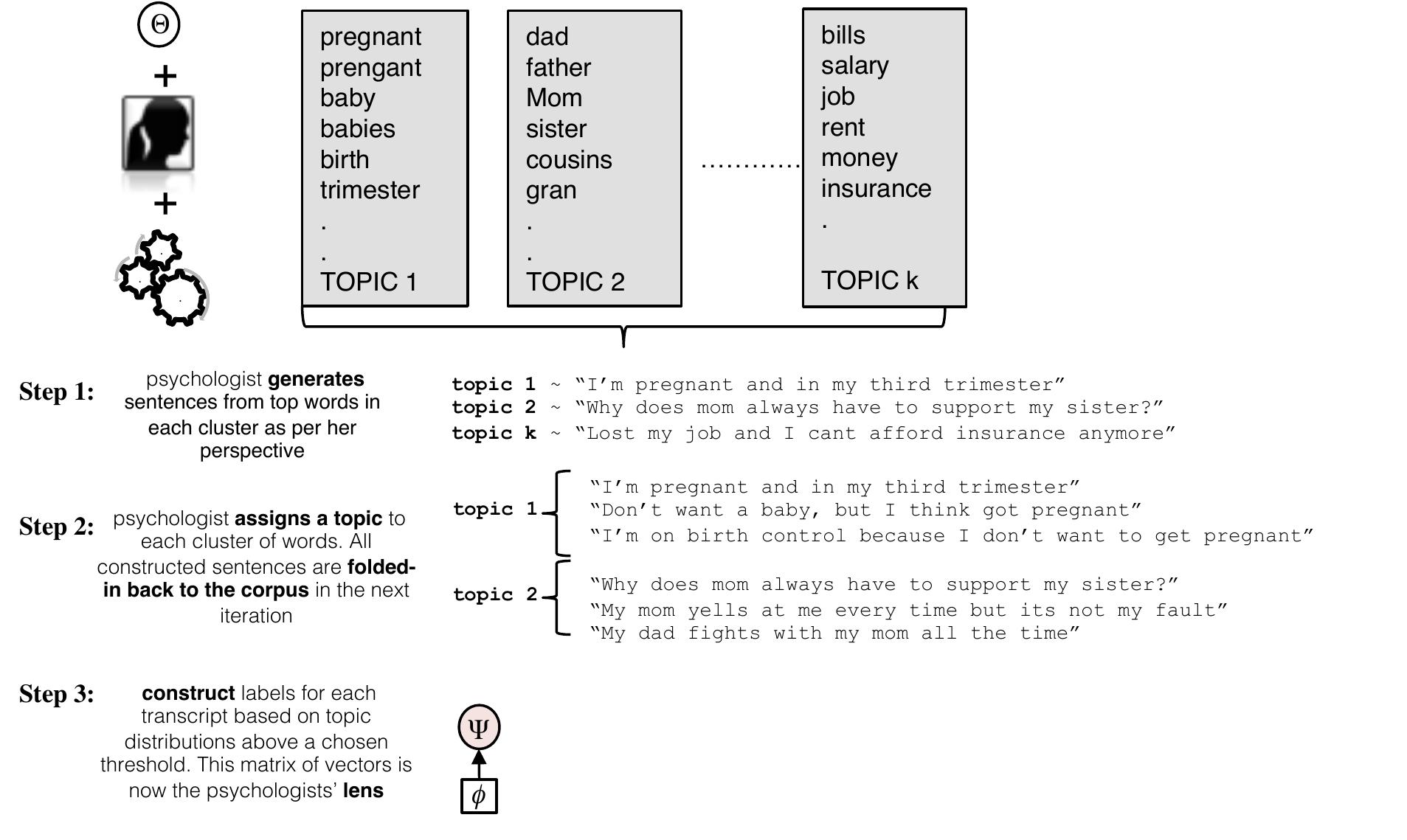}
\end{center}
\caption{\textbf{Cognitive task for lens extraction:} In step 1, The therapist generates sentences from the top 20 words from each topic cluster. If our generative model has good topic and word coherence, the inverse generative process - of generating meaningful sentences from each topic should make sense. Here we show sentences constructed from topics 1, 2 and k. In step 2, a label is assigned to the group of sentences generated for each topic. These are folded back into the corpus in the next iteration of the lensing process. in step 3,the set of labels assigned corpus-wide then automatically generates a vector of labels $\psi$ for each transcript, where topics with no labels are simply discarded. We now have a lens of the therapist that we can ingest into the generative process in the next iteration. } 
\label{fig:cognitiveTask}
\end{figure*}

\section{Lensing for mental health counseling}
Consider a corpus of transcripts between individuals requiring emotional support and therapists (who have some psychology training) providing that therapy. Assume that this interaction occurs via an online crisis counseling chat interface, where both the patient and the therapist communicate with each other using a web application. Therapists may chat simultaneously with more than one individual, increasing their cognitive overload as they move from one chat window on their web application to another. Providing a real-time, automatic summarization of each conversation through a generative model powering a visualization may help in mitigating the cognitive overload. 

Such a summarization model must extract salient topics present in each ongoing chat transcript -- with topics deemed salient as determined by therapists delivering the therapy, from their lens of training and experience. Probabilistic graphical topic models are a good candidate to train generative models for such summarization with a training corpus of $C$ thousands of chat transcripts. We encounter the following realities:

\begin{enumerate}
\item \textbf{The therapist lens is not known a priori:} The therapist lens or perspectives of salient topics worth extracting is not known and is hard to estimate directly from observed words.
\item \textbf{The goal is to maximize both topic and word coherence:} Given that our goal is to deploy a generative topic model for summarization real-time, we need good topic coherence and word coherence, both of which are notoriously hard to estimate and evaluate \citep{wallach2009evaluation}.
\item \textbf{Large scale granular annotation is expensive:} Another approach would be to treat this as a multi-class classification problem, with supervised training against annotated labels assigned by the therapist. Such an annotation process is both slow and expensive.
\end{enumerate}

\begin{figure*}[ht]
\begin{center}
\includegraphics[width=10.00\linewidth]{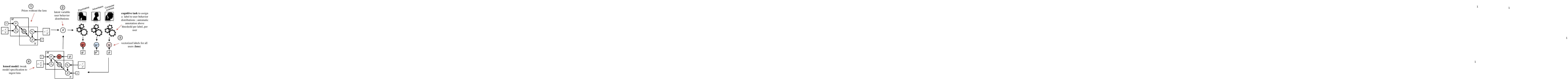}
\end{center}
\caption{\textbf{Lensing for modeling self-injurious behavior:} Readers will notice how steps 1 through 4 are similar to the ones used in \ref{fig:lensing}. The lens of the psychiatrist in unknown initially. We begin with a hierarchical Poisson matrix factorization model in step 1; the latent user behavior distributions are then presented to the psychiatrist in step 2; in step 3, the psychiatrist engages in a cognitive task where he generates complete sentences from behavior distributions to assign a semantic and salient 'theme' to each distribution. Readers should note that each user is can now be defined as distributions over these latent clusters. Some of the clusters may not make much sense and are discarded, generating a vector of behavior clusters for each user that can be thought of as the lens of the psychiatrist. There can be many lenses -- for example, if we replaced the psychiatrist with an advertiser or a computer scientist. A cluster of certain behaviors, for instance, may indicate 'negative social reinforcement' to a psychiatrist but appear non-monetizable and meaningless to an advertiser. } 
\label{fig:pmfLensing}
\end{figure*}

Let us now work our way through the lensing process for generative topic modeling of the transcripts. We choose latent dirichlet allocation, a simple yet powerful generative topic model widely used for text summarization. Readers should note the notations used here to represent LDA. $C$ is the number of transcripts in the corpus, while $N$ is the corpus-wide set of unigrams. $\alpha$, $\alpha{'}$ and $\eta$ are concentration parameter vectors of the dirichlet distribution, which are conjugate priors to the topic distributions $\theta$ and the word distributions $\beta$, which are our latent variables to be estimated after observing the words in each transcript. Each transcript is a distribution over topic distributions $\theta$. Given the parametric setup of our model, we choose $k$ number of topics. 

\begin{itemize}
\item \textbf{Step 1:} We do not have the therapist or psychologist lens
prior to the modeling process. We begin by applying LDA
with hyper-parameter optimization, as shown in step 1 of
figure 3. Hyper-parameter optimization via asymmetric
dirichlet priors has been shown to be more effective at extracting
salient topics than vanilla LDA \citep{wallach2009rethinking}.
 
\item \textbf{Step 2:} The latent topic distributions estimated are then presented to the therapist. Each topic is a distribution over corpus-wide vocabulary of size $N$. We only show the top 20 words for each topic.

\item \textbf{Step 3:} The therapist now engages in a cognitive task that takes each topic cluster of words and constructs full sentences from the each in a manner consistent with their perspective. For example, a cluster of words about family members could generate 'why does my mom always support my sister'?, referencing sibling rivalry and perceptions of thwarted belonging-ness in adolescents. If the sentences constructed from a topic cluster make sense, the therapist assigns a label to the topic. If not, the cluster is simply disregarded, as that is either not semantically meaningful (such as a collection of abbreviations such as \textit{IMO, IMHO, FYI} etc) or salient from the perspective of the therapist. In our example it the label could be 'thwarted family belonging'. The sentences so constructed are then added to the corpus for the next iteration of the lensing process. 

After the therapist has finished analyzing all the topic distributions, we now can derive labels for all transcripts with salient topics (where a transcript has a label if the topic proportion for the document is above a counselor defined threshold, say probability mass of 0.30 or above). The vector of labels for each transcript can now be thought of as a set of binarized labels $\psi_{t}$ $\sim  Bernoulli(.|\phi)$ for transcript $t$, similar to labeled LDA \citep{ramage2009labeled}. This is now our first iteration of extracting the therapist lens. $\psi$ is the therapist lens.
\item \textbf{Step 4:} We now tweak our model specification to ingest the therapist lens while simultaneously adding therapist constructed sentences (with their labels). The size of the corpus is now $C_{+}$ and the vocabulary $N_{+}$, the number of topics is now $k_{*}$ (due to the possibilities that some topics might have been discarded from the original number $k$).
\end{itemize}

Figure \ref{fig:cognitiveTask} shows an illustration of the cognitive process. Readers should note that the sentences being constructed for each topic cluster is added to our corpus in the next iteration of the lensing process, along with mediating our priors with the lens extracted from the therapist. After the therapist iterates a couple of times through the lensing steps, we arrive at a generative topic model with a good topic and word coherence. 

\subsection{Evaluation}
Evaluation of topic models for topic coherence and word coherence has been attempted intrinsically, using log-likelihood of held-out data and external corpora extrinsically such as Wikipedia. Work that has tried to revive Box's loop -- model criticism -- such as computing posterior predictive checks \citep{gelman1996posterior} via discrepancy functions \citep{mimno2011bayesian} is also another way of evaluating the models. These evaluation methods concede that statistical metrics for assessing a topic model can be challenging -- as more traditional metrics such as log-likelihood correlate negatively with others such as posterior predictive checks \citep{newman2010automatic}. While topic model evaluation continues to be an active area of research, we submit that, given the goals of embedding a therapist lens towards eventually deploying a trained model, mixed-initiative methods of evaluation make more sense \citep{dinakar2015mixed}. We discuss evaluation further after the next example of lensing. 

\section{Lensing for Modeling Self-Harm}
Consider an online social networking website for teenagers who are emotionally anguished and in need of support. Like any social network, users can post messages, comment on posts, like another user's post or comment, follow other users, and send private messages. Along with browsing information, each atomic activity a user engages in on the social network is recorded -- everything they say (in text) and click. Since the social network is to help distressed teenagers, the vast majority of posts deal with topics of depression and thoughts of self-harm, which can be annotated with a set of risk factors of mental disorders. This data can be organized into a matrix of thousands of users against hundreds of types of risk factors and online behaviors. The data is sparse -- users usually exhibit a few types of risk factors and online behaviors. For example, a user's posts might indicate a few risk factors of self-harm, categorized by past, present, or plans of engaging in self-harm. 
Predicting new user behavior by factoring large-scale behavioral data matrices from existing users is a well-studied problem in information retrieval and recommendation systems. Recent work has seen the application of probabilistic latent variable models for this purpose. One such example is hierarchical Poisson matrix factorization (HPMF), which is a powerful tool to both cluster user behaviors and for recommendation.

Let $M$ be the number of users on the social network, and let $N$ be the number of risk factors. HPMF models users as distributions over latent 'preferences' and risk factors as distribution over 'attributes'. $y_{mn}$ is the value in the matrix, $\in (0,1)$, if the $m^{th}$ user exhibited the $n^{th}$ behavior. Let $\theta^{m}$ be the latent distribution of preferences for the $m^{th}$ user, and let $\beta^{n}$ be the distribution over attributes of the '$n_{th}$  risk factor. HPMF models $y_{mn} \sim Poisson(\theta^{T}_{m}\beta_{n})$. $a,a',b',c, c',d'$ are hyperparameters to the $gamma$ distribution. Similar to the lensing example described in the previous section, the latent variable $\theta$ that defines a distribution over preferences, where a preference is a distribution over corpus-wide set of risk factors. $\theta$ can be interpreted differently by different perspectives. 
For example, a particular cluster of behaviors might indicate the vexing problem of 'social negative reinforcement (a critical function of self-harm in many people) to a psychiatrist. The same cluster might be irrelevant to an advertiser, who might be more interested in clusters of behavior that help identify users for serving them online ads. The psychiatrist and the advertiser have different goals, and different perspectives - the generative modeling for phenomenon exploration for the psychiatrist and placement of ads for the advertiser benefits from capturing their lenses and using it to inform the generative process.

The psychiatrist engages in a similar lensing process as discussed in the previous section. Readers should consult figure \ref{fig:pmfLensing} in parallel:

\begin{itemize}
\item \textbf{Step 1:} The lens of the psychiatrist, given his training in mental disorders is unknown prior to modeling. We begin by applying vanilla HPMF to the large scale sparse matrix of $M$ users against $N$ risk factors and behaviors.
\item \textbf{Step 2:} The latent distribution of user preferences over risk factors is then presented to the psychiatrist. Only the top 20 risk factors are shown for each preference.
\item \textbf{Step 3:} The psychiatrist then engages in the same cognitive task to assign semantic labels to each cluster of risk factors based on his knowledge of mental illnesses and experience treating patients. Clusters that aren't meaningful are simple ignored. As in the previous example, each user now defined as a distribution over preferences that are above a given threshold (example the preference 'social negative reinforcement' can be deemed present in a user if the probability mass is above 0.30). We now have a vector, $\psi^{'} \sim Bernoulli(0,1)$, of preferences based on the judgment of the psychiatrist. This is his lens.
\item \textbf{Step 4:} We now tweak our model specification to ingest this lens. The generative process now restricts the sampling of the latent preferences to $\psi^{'}$, respecting the judgment of the psychiatrist.
\end{itemize}

Repeating steps 2 to 4 for a 2-3 of iterations now gets us a model with user preference coherence and risk factor attribute coherence in much the same way as topic and word coherence as explained in the previous example.

\section{Evaluation}
Evaluating lensed models raises two questions: (1) how does the lensed model compare with manual annotation by the informant, and (2) how does the lensed model compare with the unlensed vanilla model?

In both applications of the lensing described above, the informant, i.e, the therapist and the psychiatrist, can manually annotate a set of transcripts and risk factors before the modeling process. Labels generated after the lensing process for these transcripts and risk behaviors can be compared with those assigned a priori before the modeling process. Conventional metrics such as ROC curves and F1-measures would indicate good performance or the lack thereof. To compare performance of the unlensed model with its lensed counterpart, the predictive labels of both models can be shown to the original informant along with the respective transcript or user risk factors. In work by Dinakar \& Lieberman et al \citep{ICWSM124604}, they compared a lensed model with a unlensed model using a Likert scale (though they did not formulate their work in terms of lensing). 

We submit that the evaluation steps described above can be a core part of model criticism advocated originally by George Box et al \citep{box2011bayesian} and recently revived by Rubin, Gelman and Blei \citep{blei2014build}. 

\section{Related Work}
There is a vast literature on user modeling, exemplified by the Springer journal \textit{User Modeling and User-Adaptive Interaction}, and the \textit{User Modeling, Adaptation, and Personalization (UMAP)} conference (umuai.org). Much of this has been directed to applications like learning user preferences for personalizing software agents \citep{konstan2012recommender}, and making student models for adaptive curricula in Intelligent Tutoring Systems \citep{desmarais2012review}.

Early work in this area did not have very sophisticated procedures for either acquiring or representing user models. For example, in Intelligent Tutoring Systems, the tutor often had a hierarchical model of the curriculum topics. Topics would be presented in order, followed by assessments to determine whether the student had mastered that curriculum topic. The student model was simply a subset of the curriculum, and the system would provide remedial tutoring for those topics not learned by the student. When the behavior of a personal agent was conditional upon a choice made by the user, the agent would observe the user's preference and cache it as a default for later use.

Webb, et.al, \citep{webb2001machine} surveyed early attempts to apply machine learning and noted the problems of required corpus size, defining concepts or topics, and complexity problems. Since then, however, tremendous advances in machine learning have mitigated many of the issues mentioned. More recently, attempts have been made to use generative machine learning models in user profiling. To cite just a couple, Fujimoto et al \citep{fujimoto2011user} used vanilla LDA to
analyze Web logs and produce visualizations of topic spaces.  Drezde and Wallach \citep{wallach2009evaluation} use LDA to create personalized models of email activities.

In previous projects, LDA or other machine learning techniques are used to learn a topic model from labeled instances in a single-shot process. There is no mixed-initiative feedback that allows comparison of the learned models to human judgment. There is no representation of viewpoint or perspective independent of the user's identity or independent of a predefined external ontology such as an educational curriculum.

Some previous work in mixed-initiative machine learning treats
classification rather than more open-ended topic modeling. Amershi et al \citep{amershi2014power} present positive and negative examples of a learned class to a user, along with reliability estimates. It allows users to drag examples from one category to another to refine the classifier.

\section{Future work}
For future work, we are looking at applying lensing to important problems in medicine -- especially cardiology, where it has been shown that differences in points of view between patients and caregivers can routinely result in misunderstandings that can lead to neglect of potentially grave patient symptoms \citep{kreatsoulas2013reconstructing}. 

We see our work as part of an emerging movement to promote human-in-the-loop machine learning \citep{amershi2014power}. Machine Learning excels in finding patterns in data. But for those patterns to be helpful to people, they have to be interpreted in ways that make sense, taking into account their viewpoints and perspectives that may not be explicitly represented in the data. If a point of view is worth 80 IQ points, as Alan Kay says, then properly representing viewpoints within a mixed-initiative process can turn a machine learning algorithm from an idiot into a genius.

\section{Acknowledgments}
We thank Dr. Rosalind Picard, Dr. Deepak Bhatt, Dr. Matthew Nock, Dr. Bob Selman, Dr. David Blei and Dr. Eric Horvitz for their incisive comments and review. Lieberman is supported by grants from DARPA and the Air Force Office of Scientific Research. 

\vspace{-0.25in}

{\parindent -10pt\leftskip 10pt\noindent
\bibliographystyle{cogsysapa}
\bibliography{bibliography}

}


\end{document}